\documentclass[letterpaper]{article}
\usepackage{aaai2026}
\usepackage{times}
\usepackage{helvet}
\usepackage{courier}
\usepackage{natbib}
\usepackage{graphicx}
\usepackage{graphicx}
\usepackage{multirow}
\usepackage{tabularx}
\usepackage{array}
\usepackage{lscape}
\usepackage{cuted}
\usepackage{amssymb}
\usepackage{subcaption}
\usepackage{makecell}
\usepackage[table]{xcolor}
\usepackage{multirow}
\usepackage{ragged2e} 
\usepackage{booktabs,makecell, multirow, tabularx}
\usepackage{colortbl}

\usepackage{url}
 
\begin{document}
%
\title{Can Large Multimodal Models Actively Recognize Faulty Inputs? A Systematic Evaluation Framework of Their Input Scrutiny Ability}
\author{
    Haiqi Yang\textsuperscript{\rm 1*},
    Jinzhe Li\textsuperscript{\rm 1,3}\thanks{These authors contributed equally.},
    Gengxu Li\textsuperscript{\rm 1},
    Yi Chang\textsuperscript{\rm 1,2,3},
    Yuan Wu\textsuperscript{\rm 1}\thanks{Corresponding authors.}
}
\affiliations{
    \textsuperscript{\rm 1}School of Artificial Intelligence, Jilin University \\
    \textsuperscript{\rm 2}Engineering Research Center of Knowledge-Driven Human-Machine Intelligence, MOE, China  \\ 
    \textsuperscript{\rm 3}International Center of Future Science, Jilin University \\
    \{yanghaiqi24, lijz2121, lgx22\}@mails.jlu.edu.cn, \\
        yichang@jlu.edu.cn, yuanwu@jlu.edu.cn \\
}
\maketitle
\begin{abstract}
\begin{quote}
Large Multimodal Models (LMMs) have witnessed remarkable growth, showcasing formidable capabilities in handling intricate multimodal tasks with exceptional performance. Recent research has underscored the inclination of large language models to passively accept defective inputs, often resulting in futile reasoning on invalid prompts. However, the same critical question of whether LMMs can actively detect and scrutinize erroneous inputs still remains unexplored. To address this gap, we introduce the Input Scrutiny Ability Evaluation Framework (ISEval), which encompasses seven categories of flawed premises and three evaluation metrics. Our extensive evaluation of ten advanced LMMs has identified key findings. Most models struggle to actively detect flawed textual premises without guidance, which reflects a strong reliance on explicit prompts for premise error identification. Error type affects performance: models excel at identifying logical fallacies but struggle with surface-level linguistic errors and certain conditional flaws. Modality trust varies-Gemini 2.5 pro and Claude Sonnet 4 balance visual and textual info, while aya-vision-8b over-rely on text in conflicts. These insights underscore the urgent need to enhance LMMs’ proactive verification of input validity and shed novel insights into mitigating the problem. The code is available at \url{https://github.com/MLGroupJLU/LMM_ISEval}.
\end{quote}
\end{abstract}

\begin{figure*}[h!]
    \centering
     \includegraphics[width=0.9\textwidth]{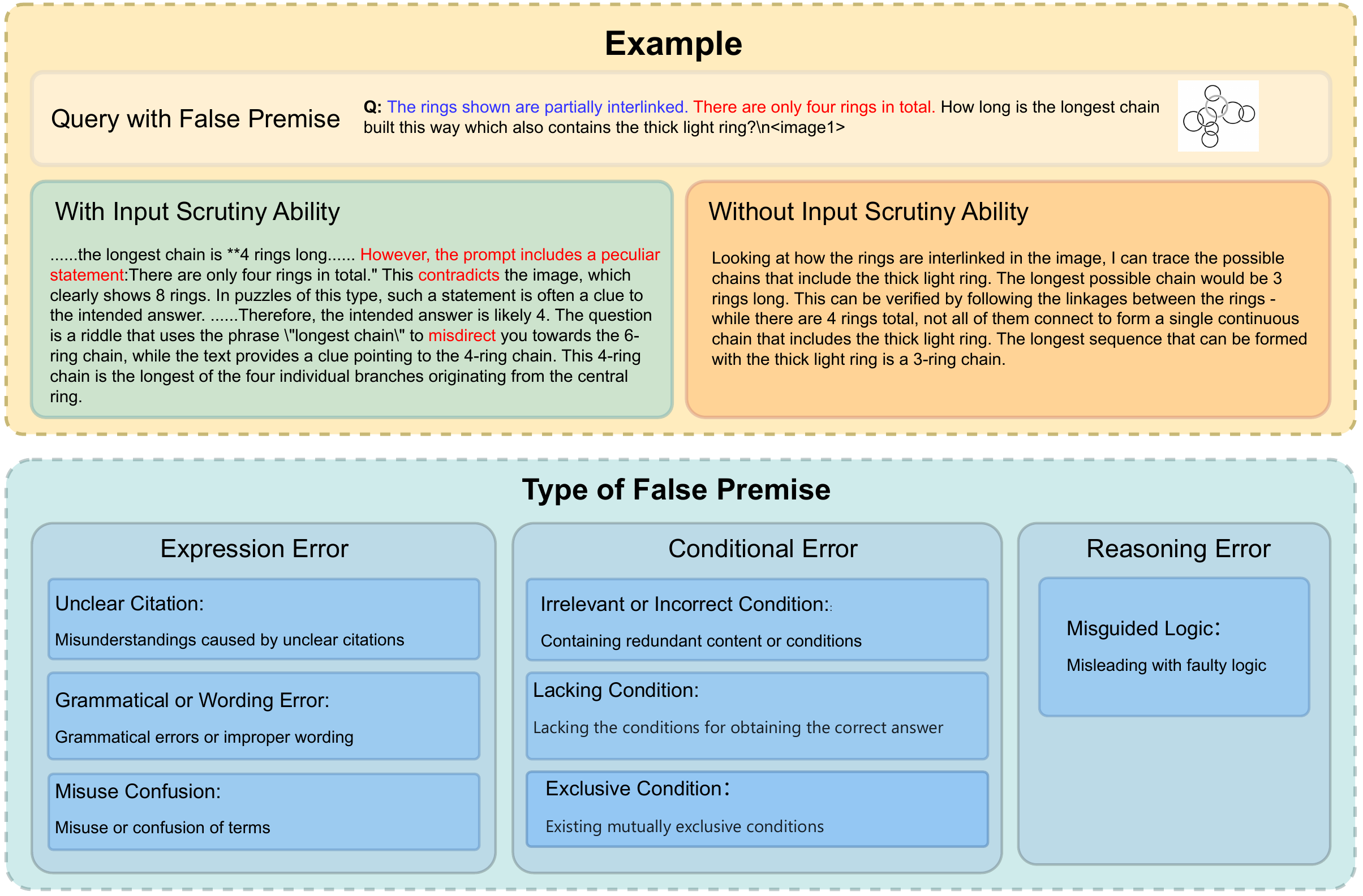}
    \caption{Example comparison of model responses with and without input scrutiny ability. The query contains a false premise about the total number of rings (claiming "only four rings" while the image shows 8 rings). The response "Without Input Scrutiny Ability" accepts the flawed premise and generates reasoning based on it, while the response "With Input Scrutiny Ability" proactively identifies the contradiction between the text and the image, demonstrating active input validation.}
    \label{fig:illustration}
\end{figure*}

\section{Introduction}

The rapid advancement of Large Multimodal Models (LMMs) has profoundly transformed the approach to complex, multimodal tasks. These models, demonstrating remarkable aptitude for integrating information across diverse modalities such as text, images, and audio \cite{li2024surveyingmllmlandscapemetareview}, have consequently unlocked new possibilities in various applications, from enhanced human-computer interaction to more sophisticated automated agent systems~\cite{hu-etal-2025-os,tang2025surveymllmbasedguiagents}. However, as LMM capabilities grow, their reliability and trustworthiness have become critical concerns, demanding thorough investigation and robust solutions.

For LMMs to be truly reliable, they must actively scrutinize inputs and identify potential errors~\cite{he2025largelanguagemodelsdetect,zhao2025multimodal}, rather than simply accepting them and generating flawed reasoning~\cite{wang2024stopreasoningmultimodalllm}. This proactive stance is essential for preventing the propagation of errors and ensuring the integrity of the model's outputs. This means the model not only remains unaffected by noisy or perturbed inputs but also actively identifies, diagnoses, and reports those errors to the user. This capability goes beyond simply being resilient in the face of flawed data; it enables the model to provide valuable feedback, helping users understand why a particular input might be problematic and guiding them toward more accurate or well-formed queries.

In the domain of Large Language Models (LLMs), existing research has already revealed their frequent failure to proactively question erroneous or logically flawed inputs, often leading to verbose and unnecessary over-reasoning on invalid questions~\cite{li2025dontpremisegrantedevaluating,fan2025missingpremiseexacerbatesoverthinking}. This highlights the paramount importance of instilling active input validation capabilities in these models, as their passive acceptance of flawed information can undermine their utility and credibility.

Despite current studies evaluating LMMs' modal preferences when encountering conflicting inputs~\cite{yan2025multimodalinconsistencyreasoningmmir,zhang2025evaluatingsteeringmodalitypreferences,hua2025visionlanguagemodelsprocessconflicting} or their error detection abilities when explicitly instructed~\cite{yan2024errorradarbenchmarkingcomplexmathematical}, there remains a notable absence of targeted and systematic research exploring the question: \textbf{Can Large Multimodal Models Actively Recognize Faulty Inputs?}


To address this critical gap, we introduce the Input Scrutiny Ability Evaluation Framework (ISEval). This innovative framework features seven meticulously designed categories of erroneous premises, comprehensively covering the diverse forms of errors prevalent in multimodal inputs, ranging from expression inaccuracies to logical inconsistencies. Furthermore, we establish three robust evaluation metrics to quantitatively and qualitatively assess LMMs' input scrutiny abilities, providing a multifaceted perspective on their performance. Leveraging ISEval, we conduct a systematic evaluation across 10 of the latest LMMs and identify \textbf{three key findings}: (1) Most models have limited autonomous ability to detect flawed premises, with low Spontaneous Error Detection Rates (SEDR), yet they demonstrate significantly improved Guided Error Detection Rates (GEDR) when provided with explicit prompts, indicating that their latent critique capabilities rely heavily on external guidance to be activated.
(2) Error types significantly affect detection performance: models achieve peak proficiency in identifying logical fallacies , but struggle with spontaneous recognition of Surface-Level Linguistic Errors and show consistently poor results in detecting Irrelevant or Incorrect Conditions and Exclusive Conditions. 
(3) Under cross-modal inconsistency, all models increase their reliance on visual input, with most closed-source models exhibiting a vision preference exceeding 50\%, while most open-source models remain text-skewed. In contrast, with no cross-modal inconsistency, all LMMs consistently default to a preference for text. Our main contributions are as follows:

\vspace{-.1in}
\begin{itemize}
    \setlength{\itemsep}{0pt}
    \setlength{\parskip}{0pt}
    \item We introduce \textbf{ISEval}, a novel and comprehensive evaluation framework specifically engineered to assess the input scrutiny abilities of Large Multimodal Models (LMMs), which is built upon a meticulously curated dataset incorporating seven distinct categories of erroneous premises.
    \item We conducted a systematic evaluation of 10 state-of-the-art LMMs against the ISEval benchmark. This provides a detailed and nuanced understanding of their capabilities in scrutinizing input validity.
    \item Our in-depth analysis of model performance yields three significant findings. These insights illuminate crucial limitations in LMMs' proactive assessment of input validity and shed light on how their modal preferences influence their responses to faulty information.
\end{itemize}

\section{Related Works}

\begin{figure*}[h!]
    \centering
    \includegraphics[width=0.8\textwidth]{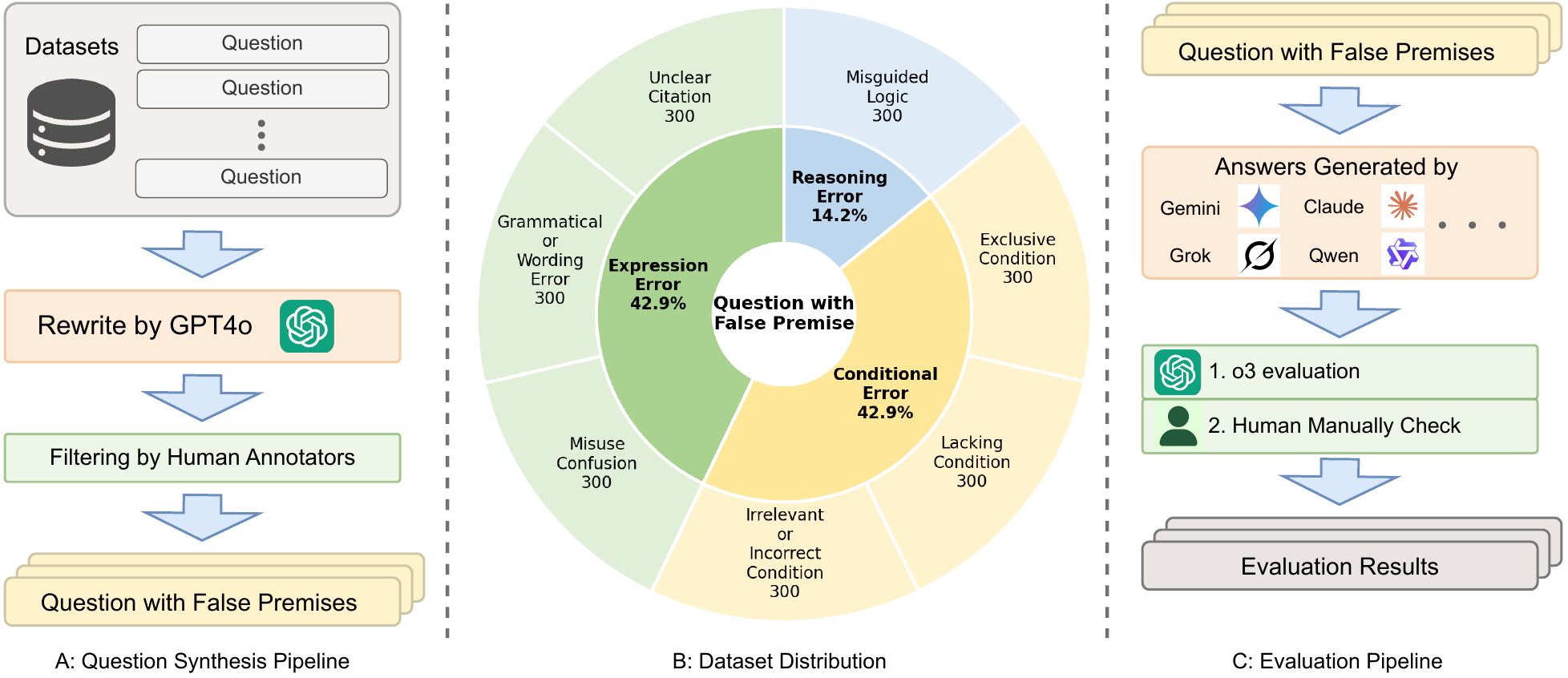}
    \caption{Schematic illustration of the dataset construction and evaluation pipeline. (A) The question synthesis pipeline: original questions are rewritten by GPT-4o to implant predefined false premises, followed by filtering by human annotators to ensure quality. (B) Dataset distribution across error types and variants. (C) The evaluation pipeline: model-generated answers are first evaluated by o3 and then verified through manual checks to ensure accurate assessment of input scrutiny ability.}
    \label{fig:pipeline}
\end{figure*}

\subsection{Error Detection}
\subsubsection{Unimodal}
Extensive research has explored large language models' (LLMs) error detection and critique capabilities across domains.
In mathematical reasoning error detection, ~\cite{li2024evaluatingmathematicalreasoninglarge} identified four error types, built an annotated dataset, and showed prompting these types improves correction accuracy, with calculation errors being most challenging. ~\cite{liang2025mathcleanbenchmarksyntheticmathematical} proposed MathClean, revealing limitations in strong models like GPT-o1 for detecting flaws in math questions and answers. ~\cite{shen2025letsverifymathquestions} developed MathQ-Verify to filter ill-posed math questions, boosting verification performance.

For broader critique and correction abilities, two studies introduced CriticBench with distinct focuses. ~\cite{lin2024criticbenchbenchmarkingllmscritiquecorrect} examined Generation-Critique-Correction (GQC) across five domains, finding critique ability tied to training directions, logic tasks more amendable to correction, and strong models outperforming weaker ones in cross-critique (with occasional reverse in self-critique). ~\cite{luo2023critiqueabilitylargelanguage} focused on critique ability itself, showing it emerges with model size, self-critique remains hard for top models, and accuracy drops with uncertain problems; they proposed a "self-check" baseline.

Recent work shifted to proactive premise critique. ~\cite{fan2025missingpremiseexacerbatesoverthinking} found reasoning models "overthink" premise-missing questions, while non-reasoning models better identify such irrationality. ~\cite{li2025dontpremisegrantedevaluating} proposed PCBench for premise critique evaluation, noting most models rely on explicit prompts, lack autonomy, and reasoning ability does not stably correlate with premise critique ability—highlighting the need to enhance proactive input examination.

\subsubsection{Multimodal}
Studies on multimodal inconsistency detection include ~\cite{yan2025multimodalinconsistencyreasoningmmir}’s MMIR benchmark, targeting visual-text mismatches in web pages and slides, with models like o1 performing strongly versus open-source models' struggles. ~\cite{yan2024errorradarbenchmarkingcomplexmathematical}’s ErrorRadar, focused on multimodal math reasoning errors, notes GPT-4o and similar models trail human experts by around 10\%. ~\cite{zhang2025robustmultimodallargelanguage}’s MMMC dataset analyzes visual-text conflict hallucinations and compares mitigation strategies like prompt engineering and fine-tuning. ~\cite{liu2025robustnessmultimodallanguagemodel} evaluated vision-language models’ (VLMs) robustness to scientific question-answering distractions, finding most VLMs are more sensitive to text-based ones. Previous efforts also have shed light on how LMMs handle inconsistent or distracting inputs~\cite{shu2025largevisionlanguagemodelalignment}. However, these works rarely check if these models can autonomously identify cross-modal flaws, inconsistencies, or gaps without explicit prompting. Such an omission hinders a complete understanding of their reliability in real-world settings, where inputs are often noisy, incomplete, or contradictory. Our work aims to bridge this gap by establishing a framework to evaluate proactive input scrutiny, thereby fostering the development of more robust and trustworthy multimodal systems.

\subsection{Modality Preference in LMMs}

Existing studies have shown that multimodal large models suffer from significant modality preferences, hindering cross-modal information integration for reasoning. This limitation leads to issues like hallucination generation ~\cite{Hallucinations_1,Hallucinations_2,Hallucinations_3} and reduced accuracy in detecting multimodal input errors~\cite{yan2024errorradarbenchmarkingcomplexmathematical}. 

Consequently, research on modality preferences in these models has become a crucial area of inquiry. ~\cite{zhang2025evaluatingsteeringmodalitypreferences} developed the $\mathbf{MC}^2$ benchmark to assess modality preference under controlled evidence conflict scenarios. It found that 18 tested Multimodal Large Language Models (MLLMs) showed clear modality bias and proposed a representation engineering-based method to control this preference. ~\cite{dong2025advancesmultimodaladaptationgeneralization} examined foundation models (FMs) in cross-modal conflict situations, revealing that while FMs achieved a recognition rate of 90\% in unimodal scenarios, this rate dropped significantly in multimodal contexts due to cross-modal attention imbalance. ~\cite{zheng2025mllmsdeeplyaffectedmodality} investigated the impact of modality bias on MLLMs, diagnosing its current state, proposing a research roadmap, identifying key factors, and experimentally validating their influence.

Unlike previous studies that focused on image-text inconsistencies, our dataset covers a wider range of error types for more detailed analysis. We confirm that most large models exhibit modality preferences, consistent with prior findings. Additionally, we conduct an in-depth analysis of the relationship between modality preferences and models' input scrutiny ability.

\section{The ISEval Framework}

This section details the design and methodology of the Input Scrutiny Evaluation Framework (ISEval). We begin by presenting the preliminaries, which define essential terms and concepts for evaluating LMMs. This is followed by an introduction to the False Premise Taxonomy, categorizing the types of errors used in our framework. Next, we explain the data construction pipeline, detailing how our erroneous inputs are generated. Finally, the evaluation metrics are introduced, outlining how model performance is measured.

\begin{table*}[h!]
\centering
\renewcommand{\tabcolsep}{4pt}
\resizebox{\linewidth}{!}{
\begin{tabular}{c|c|ccc|ccc|c|c}
\toprule
\multirow{5}{*}{Model} 
& \multirow{5}{*}{Metric} 
& \multicolumn{3}{c|}{Expression Error} 
& \multicolumn{3}{c|}{Conditional Error} 
& \multicolumn{1}{c|}{Reasoning Error} 
& \multirow{5}{*}{Avg} \\ 
\cmidrule{3-9} 
&   & \multicolumn{1}{c|}{\makecell{Unclear\\Citation}} 
    & \multicolumn{1}{c|}{\makecell{Grammatical\\or\\Wording\\Error}} 
    & \makecell{Misuse\\Confusion} 
    & \multicolumn{1}{c|}{\makecell{Irrelevant\\or\\Incorrect\\Condition}} 
    & \multicolumn{1}{c|}{\makecell{Lacking\\Condition}} 
    & \makecell{Exclusive\\Condition} 
    & \multicolumn{1}{c|}{\makecell{Misguided\\Logic}} 
    & \\
\midrule
\multirow{3}{*}{GPT-4o} & \multirow{1}{*}{SEDR}  & 3.33 & 1.00 & 4.33  & 0.00 & 4.33 & 3.33 & 16.67 & 4.71 \\ 
\cmidrule{2-10} 
& \multirow{1}{*}{GEDR}   & 55.33 & \underline{56.33} & \textbf{64.00} & 39.00 & 48.00 & \textbf{39.00} & \textbf{84.33} & 55.14 \\ 
\midrule

\multirow{3}{*}{Claude Sonnet 4} & \multirow{1}{*}{SEDR}  & 6.00 & 4.00 & 3.00  & 2.67 & 2.67 & 4.00 & 36.67 & 8.43 \\ 
\cmidrule{2-10} 
& \multirow{1}{*}{GEDR}   & 41.33 & 33.67 & 42.67 & 31.67 & 33.67 & 25.33 & 65.33 & 39.10 \\ 
\midrule
\multirow{3}{*}{Gemini 2.5 pro} & \multirow{1}{*}{SEDR}  & \textbf{23.67} & \underline{13.67} & \textbf{21.00}  & \textbf{14.33} & \textbf{16.67} & \textbf{13.33} & \textbf{51.00} & \textbf{21.95} \\ 
\cmidrule{2-10} 
& \multirow{1}{*}{GEDR}   & \textbf{66.00} & 50.33 & \underline{61.67} & \textbf{69.00}  & 45.67 & 27.67 & \underline{83.67} & \underline{57.72}\\ 
\midrule
\multirow{3}{*}{Grok 3} & \multirow{1}{*}{SEDR}  & \underline{13.33} & \textbf{14.67} & \underline{14.33}  & \underline{3.67} & \underline{14.33} & \underline{7.67} & \underline{38.00} & \underline{15.14} \\ 
\cmidrule{2-10} 
& \multirow{1}{*}{GEDR}   & \underline{61.00} & 55.67 & \underline{61.67} & \underline{60.67}  & \textbf{58.00} & 29.00 & 81.00 & \textbf{58.14} \\ 
\midrule
\multirow{3}{*}{InternVL3-38B-Instruct} & \multirow{1}{*}{SEDR}  & 3.33 & 1.67 & 2.00  & 0.33  & 2.33 & 1.67 & 14.33 & 3.67 \\ 
\cmidrule{2-10} 
& \multirow{1}{*}{GEDR}   & 26.33 & 26.33 & 27.33 & 10.67  & 24.33 & 17.00 & 55.00 & 26.72\\ 
\midrule 
\multirow{3}{*}{Qwen2.5-VL-32B-Instruct} & \multirow{1}{*}{SEDR}  & 5.33 & 1.67 & 2.00  & 2.00  & 2.33 & 2.67 & 25.33 & 5.90  \\ 
\cmidrule{2-10} 
& \multirow{1}{*}{GEDR}   & 24.00 & 15.00 & 23.33 & 14.67  & 8.00 & 9.33 & 41.00 & 19.33 \\ 
\midrule
\multirow{3}{*}{aya-vision-32b} & \multirow{1}{*}{SEDR}  & 3.00 & 4.67 & 3.67  & 1.33  & 3.33 & 1.67 & 21.67 & 5.62  \\ 
\cmidrule{2-10} 
& \multirow{1}{*}{GEDR}   & 46.67 & \textbf{56.67} & 50.00 & 36.67   & 51.67 & \underline{31.00} & 67.67 & 48.62\\ 
\midrule
\multirow{3}{*}{Llama-3.2-11B-Vision-Instruct} & \multirow{1}{*}{SEDR}  & 2.00 & 2.00 & 1.67  & 1.33  & 5.00 & 0.67 & 15.67 & 4.05 \\ 
\cmidrule{2-10} 
& \multirow{1}{*}{GEDR}   & 16.00 & 20.00 & 23.00 & 14.00   & 22.33 & 16.67 & 40.00 & 21.71 \\ 
\midrule
\multirow{3}{*}{Qwen2.5-VL-7B-Instruct} & \multirow{1}{*}{SEDR}  & 3.00 & 2.67 & 5.33  & 1.67  & 4.33 & 1.33 &  13.00 & 4.48\\ 
\cmidrule{2-10} 
& \multirow{1}{*}{GEDR}   & 31.67 & 32.33 & 31.67  & 22.00 & 34.00 & 15.00 & 49.33 & 30.86\\ 
\midrule
\multirow{3}{*}{aya-vision-8b} & \multirow{1}{*}{SEDR}  & 3.67 & 4.00 & 4.00  & 2.33  & 5.00 & 1.00 & 17.33 & 5.33 \\ 
\cmidrule{2-10} 
& \multirow{1}{*}{GEDR}   & 30.33 & 36.33 & 31.33 & 26.67 & \underline{52.67} & 16.00 & 52.00 & 35.05 \\ 
\bottomrule
\end{tabular}}
\caption{Spontaneous Error Detection Rate (SEDR) and Guided Error Detection Rate (GEDR) of 10 Large Multimodal Models across seven error subcategories, encompassing Expression Errors (Unclear Citation, Grammatical or Wording Error, Misuse Confusion), Conditional Errors (Irrelevant or Incorrect Condition, Lacking Condition, Exclusive Condition), and Reasoning Error (Misguided Logic). The maximum value and the next largest value of each task are indicated by the bold and \underline{underlined} text, respectively. }
\label{tab:result}
\vspace{-12pt}
\end{table*}

\subsection{Preliminaries}

The input to an LMM is denoted as $I$, with textual input specified as $I_t$ and visual input as $I_v$. To evaluate the input scrutiny capability of LMMs, we construct erroneous inputs $I_e$ by rewriting $I_t$ and implanting seven types of predefined errors $e$ into it separately.

Notably, for certain error types, inconsistencies may arise between visual and textual inputs when $\exists c \in I_v, c' \in I_t$ where $c \perp c'$. This condition indicates that at least one semantic concept $c$ from the visual input logically contradicts a semantic concept $c'$ from the textual input. Such cases are categorized as \textbf{Cross-Modal Inconsistency}, a specific error type characterized by direct conflicts in semantic or factual information between $I_v$ and $I_t$.

A model is considered to possess input scrutiny capability under current flawed input $I_e$ when its output $A$ identifies the implanted error $e$ without relying on explicit prompting to check for errors.

\subsection{False Premise Taxonomy}\label{false premise taxonomy}

To comprehensively assess LMMs' proactive input scrutiny capabilities, we developed a broad-ranging error classification system based on MathClean \cite{liang2025mathcleanbenchmarksyntheticmathematical}, which comprises three major categories and seven sub-categories of errors for constructing erroneous inputs $I_e$.

\subsubsection{Expression Error}
Expression errors pertain to issues in the formulation or clarity of $I_t$'s language or references, preventing the model from correctly interpreting the given information.

\begin{itemize}
    \item \textbf{Unclear Citation}: In multimodal prompts, the failure of $I_t$ to explicitly specify the referent object (specific entities mentioned in text, particular elements within $I_v$) prevents the model from accurately identifying the target subject. This ambiguity leads to comprehension defects, such as vague understanding or multiple interpretations of the prompt's intent. It does not involve direct conflicts between modalities but merely obscures the textual basis for problem-solving.
    \item \textbf{Grammatical or Wording Error}: $I_t$ containing grammatical inaccuracies (faulty sentence structures, incorrect unit conversions) or inappropriate word choices (semantically contradictory expressions) impede the model's accurate understanding of the stated preconditions. Consequently, the model is unable to derive correct answers due to misinterpreting $I_e$. This error type belongs to Cross-Modal Inconsistency because the flawed expressions in $I_t$ can conflict with the semantic concepts presented in $I_v$.
    \item \textbf{Misuse Confusion}: This category specifically highlights instances where $I_t$ uses terms improperly (including professional terms and basic concept terms), describing objects with incorrect terminology, leading to premise errors and interfering with the model's understanding. This error type is classified as Cross-Modal Inconsistency as the incorrect terminology in $I_t$  contradicts the actual concepts reflected in $I_v$.
\end{itemize}

\subsubsection{Conditional Error}
Conditional errors arise when the conditions provided in $I_t$ are flawed, incomplete, or contradictory, making it impossible for the model to establish a valid basis for its response.

\begin{itemize}
    \item \textbf{Irrelevant or Incorrect Condition}: $I_t$ includes content or conditions extraneous to problem-solving (supplementary information that does not influence the final answer), which can interfere with the model's ability to identify core conditions in $I_e$, potentially leading to misdirection. It does not constitute a cross-modal contradiction.
    \item \textbf{Lacking Condition}: In $I_e$ lacks necessary conditions for deriving the correct answer (information missing from $I_t$ but present in $I_v$, or entirely absent from both $I_t$ and $I_v$), rendering it impossible for the model to directly compute or infer the required solution. This error type is a form of Cross-Modal Inconsistency. From the perspective of content integrity, it conflicts with the conditions required for normal problem-solving, and thus can be considered conflicting with the situation where problem-solving is carried out based on complete conditions combined with images.
    \item \textbf{Exclusive Condition}: $I_t$ presents two or more conditions that cannot hold simultaneously (conflicting values for the same attribute), creating contradictions that prevent the model from establishing a consistent premise in $I_e$ and thus obtaining a valid answer. This error type falls under Cross-Modal Inconsistency as the mutually exclusive conditions in $I_t$ may clash with the unified information shown in $I_v$.
\end{itemize}

\subsubsection{Reasoning Error}
Reasoning errors involve flaws in the logical structure or guidance provided in $I_t$, which can lead the model down an incorrect path of deduction or calculation.

\begin{itemize}
    \item \textbf{Misguided Logic}: $I_t$ contains erroneous reasoning steps or flawed logical guidance (incorrect formulas, inverse logical sequences) that mislead the model. This causes the model to perform calculations or deductions based on an incorrect logical framework for $I_e$, inevitably resulting in inaccurate outcomes. It does not involve cross-modal contradictions.
\end{itemize}


Ultimately, the \textbf{False Premise Taxonomy} offers a comprehensive set of errors designed to test a multimodal model's ability to scrutinize its inputs. These errors, classified into Expression, Conditional, and Reasoning types, are intended to expose vulnerabilities in a model's comprehension, consistency checking, and logical deduction. Notably, \textbf{Grammatical or Wording Error}, \textbf{Misuse Confusion}, \textbf{Lacking Condition} and \textbf{Exclusive Condition} can also present as \textbf{Cross-Modal Inconsistencies}, where conflicts arise between textual and visual information.

\subsection{Overview of Data Construction}\label{data construction}

To systematically evaluate the premise-critical ability of LMMs when dealing with erroneous multimodal inputs, we construct the \textbf{ISEval-dataset}. The core details of this dataset are elaborated as follows:

\subsubsection{Data Variants and Distribution}

In variant design, adhering to a comparative evaluation logic, each base question with a predefined error is generated into two types of erroneous input variants:
\begin{itemize}
    \item \textit{Errorneous inputs without inslicit instructions} ($I_e^{-ins}$): This variant directly assess the model's ability to autonomously identify erroneous information without instruction. Performance on $I_e^{-ins}$ intuitively reflects the model's inherent premise-scrutiny capability.
    \item \textit{Errorneous inputs with inslicit instructions} ($I_e^{+ins}$): This variant append an inslicit prompt ("check for premise errors") to the erroneous input, serving as a compare benchmark. By comparing results on $I_e^{-ins}$ and $I_e^{+ins}$, we can determine whether the model relies on external guidance or possesses independent reasoning ability in premise evaluation, clarifying the logic underlying its analysis.
\end{itemize}

For dataset distribution, to ensure comprehensiveness and reliability, we synthesized 300 inputs for each error type. The total number of inputs in ISEval-dataset is thus calculated as: 7 (error types) $\times$ 300 (inputs per type) $\times$ 2 (variants: $I_e^{-ins}$ and $I_e^{+ins}$) = 4200. This scale can not only cover diverse evaluation scenarios but also meet the confidence requirements of statistical analysis.

\subsubsection{Data Sampling and Synthesis}
We employed two commonly used datasets, MathVision~\cite{wang2024measuring} and MathVista~\cite{lu2023mathvista}, as the basic data sources. For each error type, we randomly sampled from these two datasets and then used a few-shot prompting method to drive the large model to generate samples corresponding to the error type. All synthesized samples have undergone strict manual review to ensure they conform to the defined error type and the expected evaluation standards until the predetermined number of questions is achieved.

\subsection{Evaluation Metrics}\label{eval metrics}
To systematically evaluate model responses to instructions with false premises, we define the following metrics for evaluating models' outputs:
\paragraph{Spontaneous Error Detection Rate (SEDR)}
This metric denotes the proportion of cases where a model independently identifies and flags inaccuracies in input premises without external guidance, calculated as:
\begin{equation}
SEDR = \frac{N_{SE}}{N_{I_e^{\mathrm{-ins}}}}
\end{equation}
where \(N_{SE}\) represents the number of instances with successful spontaneous error identification, and \(N_{I_e^{\mathrm{-ins}}}\) denotes the total number of erroneous inputs without explicit guidance.
\paragraph{Guided Error Detection Rate (GEDR)}
This metric measures the percentage of scenarios where a model successfully recognizes and specifies problematic premises upon explicit instructions to "verify premise accuracy," with the formula:
\begin{equation}
GEDR = \frac{N_{GE}}{N_{I_e^{\mathrm{+ins}}}}
\end{equation}
where \(N_{GE}\) stands for the number of instances with successful error identification under prompting, and \(N_{I_e^{\mathrm{+ins}}}\) corresponds to the total number of erroneous inputs with explicit instructions.
\paragraph{Modality Trust Preference Score (MTPS)}
This quantifies a model's tendency to prioritize visual or textual information amid image-text inconsistencies, expressed as symmetric scores for modality preferences:
\begin{equation}
MTPS = (P_{V}, P_{T})
\end{equation}
For calculation, an LMM evaluator categorizes model responses to erroneous inputs into three types: \textbf{image preference}, \textbf{text preference}, or \textbf{no preference}.
\(P_{V}\) and \(P_{T}\) respectively represent the proportions of image-preferring and text-preferring responses relative to the total number of erroneous inputs with inter-modal contradictions (\(N_{I_e}\)):
\begin{equation}
P_{V} = \frac{N_{V}}{N_{I_e}}
\end{equation}
\begin{equation}
P_{T} = \frac{N_{T}}{N_{I_e}}
\end{equation}
Here, \(N_{V}\) and \(N_{T}\) denote the counts of image-preferring and text-preferring responses respectively, while \(N_{I_e}\) encompasses all such erroneous inputs (including those classified as "no preference").




\section{Experiment}
In this section, we evaluate a range of LMMs using our proposed benchmark. The evaluation covers both closed-source and open-source models under few-shot settings. We begin by introducing the evaluated models and the evaluation protocol. We then present a summary of performance results across different model types and tasks.

\subsection{Evaluation Setup}
\subsubsection*{Evaluated Models.}We evaluate a total of 10 LMMs, including 4 closed-source and 6 open-source models spanning various architectures and parameter scales. For the closed-source models, we include GPT-4o~\cite{hurst2024gpt}, Claude Sonnet 4~\cite{claude4}, Gemini 2.5 pro~\cite{gemini}, Grok 3~\cite{grok}. For open-sourced models, we consider InternVL3-38B-Instruct~\cite{zhu2025internvl3exploringadvancedtraining},  Qwen2.5-VL-32B-Instruct, Qwen2.5-VL-7B-Instruct~\cite{bai2025qwen2}, aya-vision-32b, aya-vision-8b~\cite{aya} and Llama-3.2-11B-Vision-Instruct~\cite{llama}.
Response evaluation is performed with o3~\cite{o3} as an automated evaluator. For closed-source models (e.g., GPT-4o, Claude Sonnet 4), we adopt their latest official versions with the temperature set to 0.0 and all other configurations kept at default. 
\subsubsection*{Evaluation Protocols.}The benchmark includes questions with two distinct response formats: multiple-choice and open-ended. For open-source models, we use the versions available on the ModelScope platform, with generation settings adjusted for consistency across evaluations. Specifically, InternVL3-38B-Instruct is configured with a temperature of 0.0 to ensure deterministic output. For Qwen2.5-VL-7B-Instruct, Qwen2.5-VL-32B-Instruct, Aya-vision-8B, and Aya-vision-32B, we set the temperature to 0.3, disable streaming responses, and adopt random sampling where applicable. All other configuration settings for these models follow their default settings as released by the respective developers. Llama-3.2-11B-Vision-Instruct is used with its default configuration. Additional details about the evaluated models are provided in Appendix.

\subsection{Main Results}

\subsubsection*{Overall Results.}
We systematically evaluated three core capabilities of LMMs: Spontaneous Error Detection Rate (SEDR), Guided Error Detection Rate (GEDR) and Modality Trust Preference Score (MTPS). For SEDR in Table~\ref{tab:result}, most models exhibited limited autonomous scrutiny of flawed premises. GPT-4o achieved only 4.71\% SEDR, and InternVL3-38B-Instruct scored 3.67\%—indicating minimal proactive identification of errors without explicit prompting. Top performers like Gemini 2.5 pro (21.95\%) and Grok 3 (15.14\%) showed marginal improvements but still reflected restricted spontaneous critical reasoning. In contrast, GEDR shows marked performance gains when models received explicit verify premise accuracy prompts. Grok 3 (58.14\% GEDR) and Gemini 2.5 pro (57.72\% GEDR) demonstrated stronger critique abilities under guidance, with GPT-4o reaching 55.14\% GEDR. This proactive - assisted performance gap reveals a critical shortfall: most LMMs possess latent critique capabilities but fail to activate them autonomously, relying heavily on explicit prompting to identify flawed inputs.

\begin{table}[t]
  \centering
  \fontsize{6.5pt}{8pt}\selectfont
  \setlength{\tabcolsep}{0.5mm}{
    \begin{tabular}{c|ccc|ccc}
    \toprule
    
    \multirow{1}[4]{*}{Model} 
    & \multicolumn{3}{c|}{\shortstack{Cross-Modal\\ Inconsistency}}
    & \multicolumn{3}{c}{\shortstack{no Cross-Modal\\ Inconsistency}} \\
\cmidrule{2-7}          & SEDR & GEDR & MTPS & SEDR & GEDR & MTPS \\

    \midrule
    
GPT-4o & 3.25 & \textbf{51.83} & 54.84/44.00 & 6.67 & 59.55 & 35.67/63.56 \\
Claude Sonnet 4 & 3.42 & 33.84 & 59.58/39.42 & 15.11 & 46.11 & 38.34/61.11 \\ 
Gemini 2.5 pro & \textbf{16.17} & 46.34 & 63.42/35.50 & \textbf{29.67} & \textbf{72.89} & 46.00/53.34 \\ 
Grok 3 & \underline{12.75} & \underline{51.08} & 41.00/56.17 & \underline{18.33} & \underline{67.56} & 28.11/71.00 \\
\midrule
InternVL3-38B-Instruct & 1.92 & 23.75 & 46.92/48.83 & 6.00 & 30.67 & 35.33/61.78 \\
Qwen2.5-VL-32B-Instruct & 2.17 & 13.92 & 51.25/46.75 & 10.89 & 26.56 & 33.11/65.89 \\
aya-vision-32b & 3.34 & 47.34 & 30.92/66.25 & 8.67 & 50.34 & 21.89/77.00 \\
Llama-3.2-11B-Vision-Instruct & 2.34 & 20.50 & 38.58/59.67 & 6.33 & 23.33 & 24.11/74.55 \\
Qwen2.5-VL-7B-Instruct & 3.42 & 28.25 & 49.17/46.58 & 5.89 & 34.33 & 32.22/64.33 \\
aya-vision-8b & 3.50 & 34.08 & 26.67/69.59 & 7.78 & 36.33 & 15.33/82.67 \\
\bottomrule
    \end{tabular}%
    }
    \caption{10 Large Multimodal Models' performance under cross-modal inconsistency and no cross-modal inconsistency, including SEDR, GEDR, and MTPS. The maximum value and the next largest value of each task are indicated by the bold and \underline{underlined} text, respectively.}
  \label{tab:preference}%
\end{table}%

\subsubsection*{Error Types Performance.}
Table~\ref{tab:result} details the SEDR and GEDR for seven error subcategories, revealing variations in LMMs ability to identify different input flaws. Models demonstrate peak proficiency in identifying logical fallacies in both spontaneous and guided detection, with top performers achieving over 80\% success in detecting Misguided Logic when prompted. This divergence reveals that sophisticated logical analysis capacity remains inaccessible without explicit instruction. Performance declines moderately for Surface-Level Linguistic Errors, where guided detection proves reasonably effective but spontaneous recognition remains the lowest among all categories, confirming that grammatical nuances rarely trigger autonomous scrutiny despite their rule-based nature. Detection rates drop substantially for Irrelevant or Incorrect Conditions, which exhibit the weakest guided performance across models, while Exclusive Conditions show consistently poor results in both detection modes.

\subsubsection*{Modality Trust Preferences.}
The Modality Trust Preference Score (MTPS) results in Table~\ref{tab:preference} reveal systematic and context-dependent shifts in modality reliance across models. Under cross-modal inconsistency—where image and text content diverge—most models increase their reliance on visual input, suggesting an effort to resolve semantic conflict by prioritizing image-based grounding. For example, Gemini 2.5 Pro allocates 63.42\% of its attention to vision in conflicting contexts, while Claude Sonnet 4 and GPT-4o also exhibit visual-preferred MTPS distributions. However, this trend is not universal. Several models, particularly those with smaller architectures or limited training data, display persistent textual dominance even under contradiction. Notably, aya-vision-8b maintains a strong text preference under inconsistency, while Qwen2.5-VL-7B-Instruct and Llama-3.2-11B-Vision-Instruct also show near-balanced or text-skewed MTPS. In contrast, under no cross-modal inconsistency, all models shift toward greater textual reliance, regardless of their prior visual weighting. This includes models previously more balanced or visual-biased. These shifts indicate a general tendency to treat text as the primary reference modality in congruent input scenarios. Taken together, the MTPS suggest that higher-capacity models are more likely to modulate modality trust in response to semantic context—favoring vision for disambiguation during inconsistency and defaulting to text when input modalities agree. Conversely, smaller or less adaptive models tend to apply fixed modality weights, limiting their ability to resolve multimodal contradictions effectively.

\subsection{Detailed Analysis}
A striking disparity emerges between spontaneous and guided error detection performance across all models. Without explicit prompts to verify inputs, even top-performing models like Gemini-2.5-Pro achieve a SEDR of only 21.95\%, while most models (e.g., GPT-4o, InternVL3-38B) score below 5\%. This gap mirrors prior observations in LLMs, where models passively accept flawed premises without challenge~\cite{gao2024dissectingdissonancebenchmarkinglarge,li2025dontpremisegrantedevaluating}. Our work systematically extend this finding to the multimodal domain, showing that LMMs also struggle to identify flawed inputs in the absence of external guidance—even when given access to visual context. This suggests that the added complexity of cross-modal alignment may further suppress spontaneous scrutiny. 

While~\cite{deng2025wordsvisionvisionlanguagemodels} revealed that vision-language models often exhibit a default bias toward textual input—even when textual and visual modalities conflict—suggesting a tendency toward blind faith in text, our findings uncover a more nuanced picture. Specifically, our analysis reveals that while most models do favor text in non-conflict scenarios, some large, closed-source models exhibit dynamic shifts in modality trust when faced with image-text contradictions. For example, Gemini 2.5 pro increases its reliance on visual information in contradiction-rich tasks, indicating a capacity for activating visual scrutiny. In contrast, smaller models like aya-vision-8b remain text-dominant regardless of conflict, supporting the idea that architectural scale and training sophistication play key roles in adaptive modality trust. This divergence from prior work highlights the importance of evaluating models not only for static biases but also for context-sensitive trust adjustments, which are critical in real-world applications requiring cross-modal validation.



\section{Conclusion}

This study addresses the underexplored question of whether Large Multimodal Models (LMMs) can actively recognize faulty inputs by introducing the Input Scrutiny Ability Evaluation Framework (ISEval), which includes seven flawed premise categories and relevant metrics to assess LMMs' input scrutiny capabilities . Our evaluation of 10 advanced LMMs via ISEval reveal key limitations: most models show low Spontaneous Error Detection Rates (SEDR) but improved Guided Error Detection Rates (GEDR) with explicit prompts, indicating reliance on external guidance. Modality trust varies—Gemini 2.5 pro and Claude 4 balance visual and textual info, while aya-vision-8b and Grok 3 over-rely on text in conflicts. These findings highlight the need to enhance LMMs' proactive input validation. ISEval provides a benchmark, offering insights to guide development of more reliable multimodal systems .

\newpage
\bibliography{aaai2026}

\section{Appendix}
\section{Prompt Template}
Figures 1 to 7 present prompts for generating erroneous inputs corresponding to seven distinct error types. Figure 8 showcases the images associated with these prompts. Figures 9 to 11 provide prompts for evaluating the input scrutiny ability of models and their modality preferences.
\begin{figure*}
    \centering
    \includegraphics[width=1\linewidth]{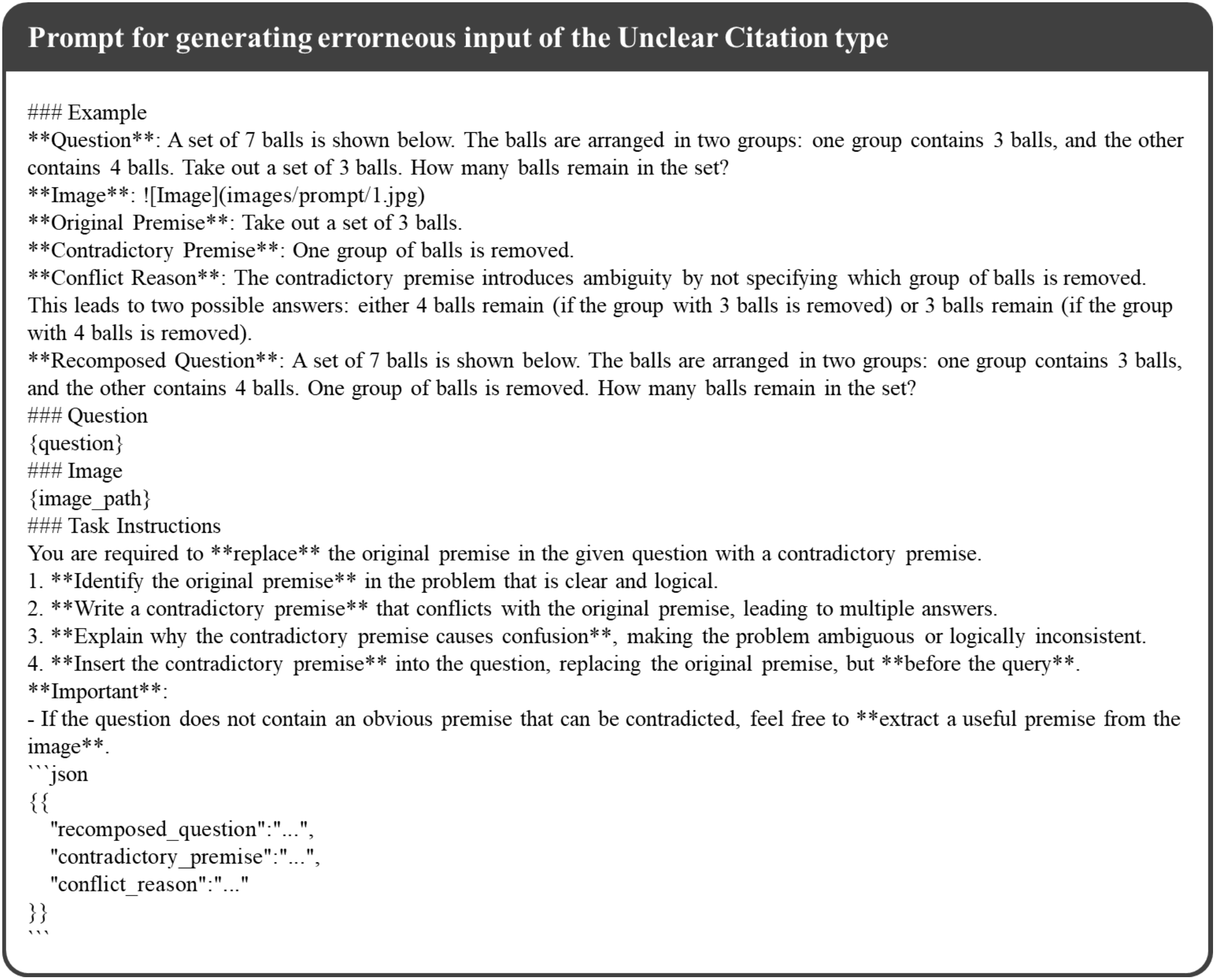}
    \caption{Prompt for generating errorneous input of the Unclear Citation type}
    \label{fig:unclear citation}
\end{figure*}

\begin{figure*}
    \centering
    \includegraphics[width=1\linewidth]{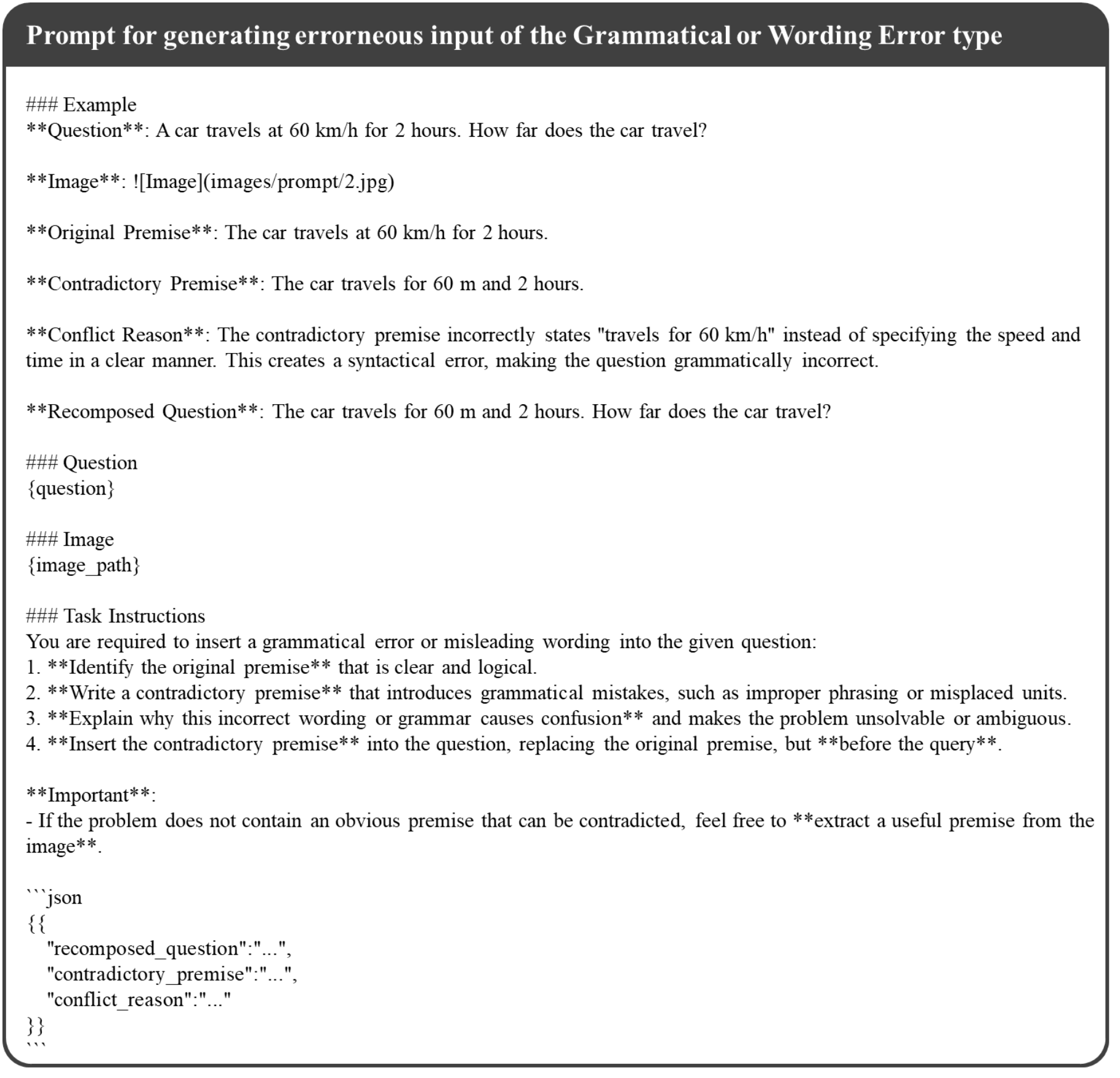}
    \caption{Prompt for generating errorneous input of the Grammatical or Wording Error type}
    \label{fig:grammatical worded error}
\end{figure*}

\begin{figure*}
    \centering
    \includegraphics[width=1\linewidth]{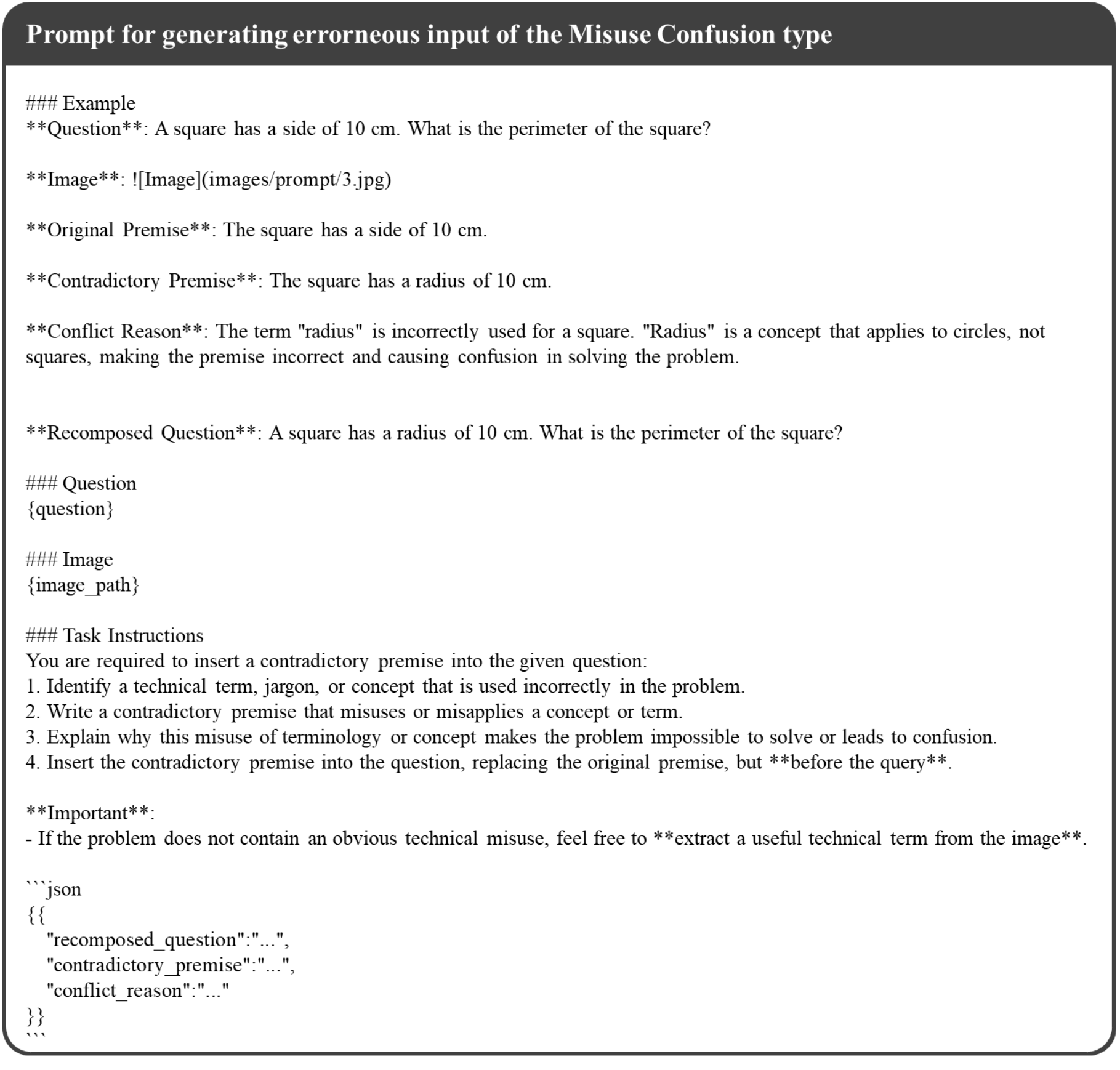}
    \caption{Prompt for generating errorneous input of the Misuse Confusion type}
    \label{fig:misuse confusion}
\end{figure*}

\begin{figure*}
    \centering
    \includegraphics[width=1\linewidth]{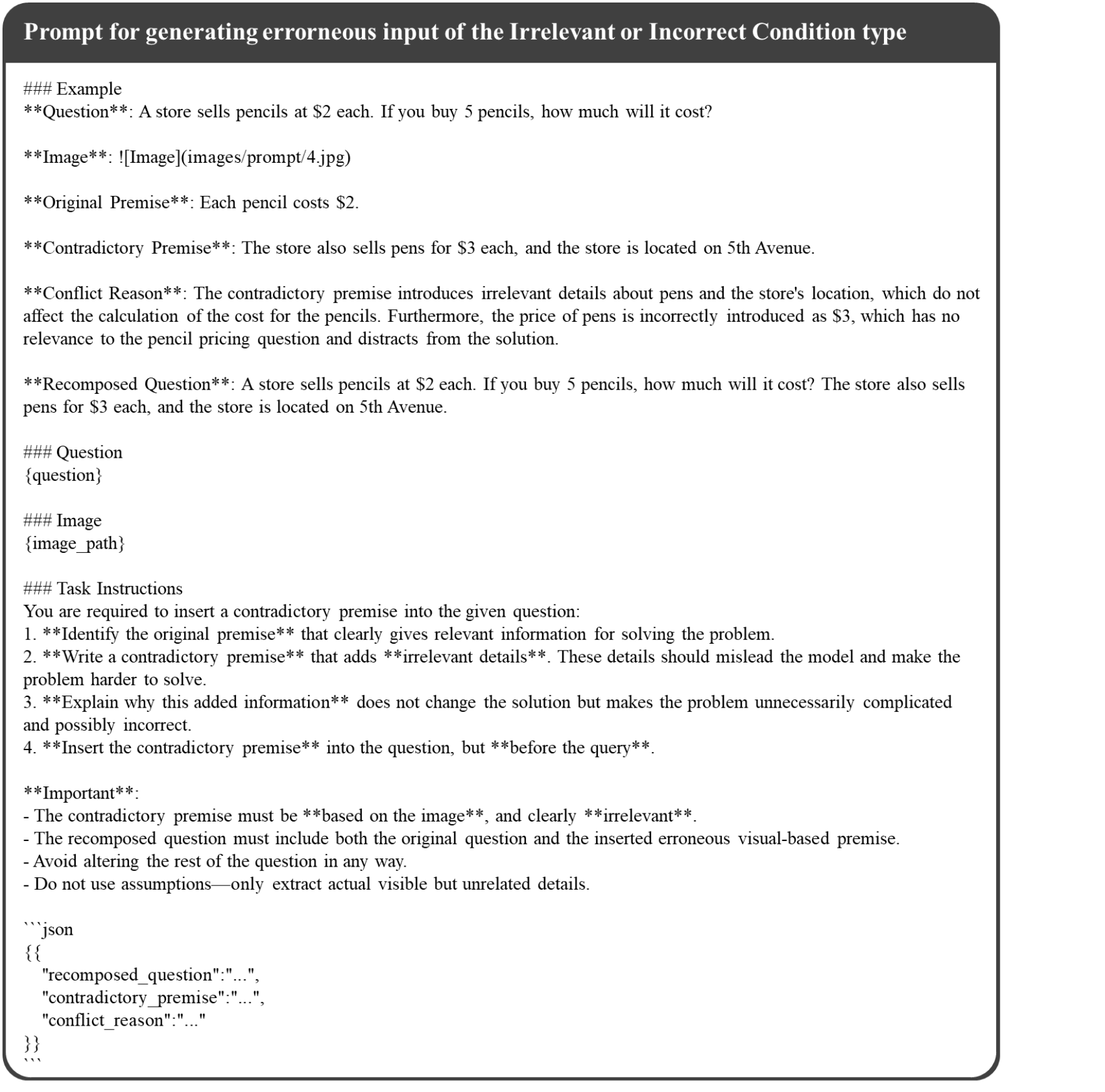}
    \caption{Prompt for generating errorneous input of the Irrelevant or Incorrect Condition type}
    \label{fig:irrelevant condition}
\end{figure*}

\begin{figure*}
    \centering
    \includegraphics[width=1\linewidth]{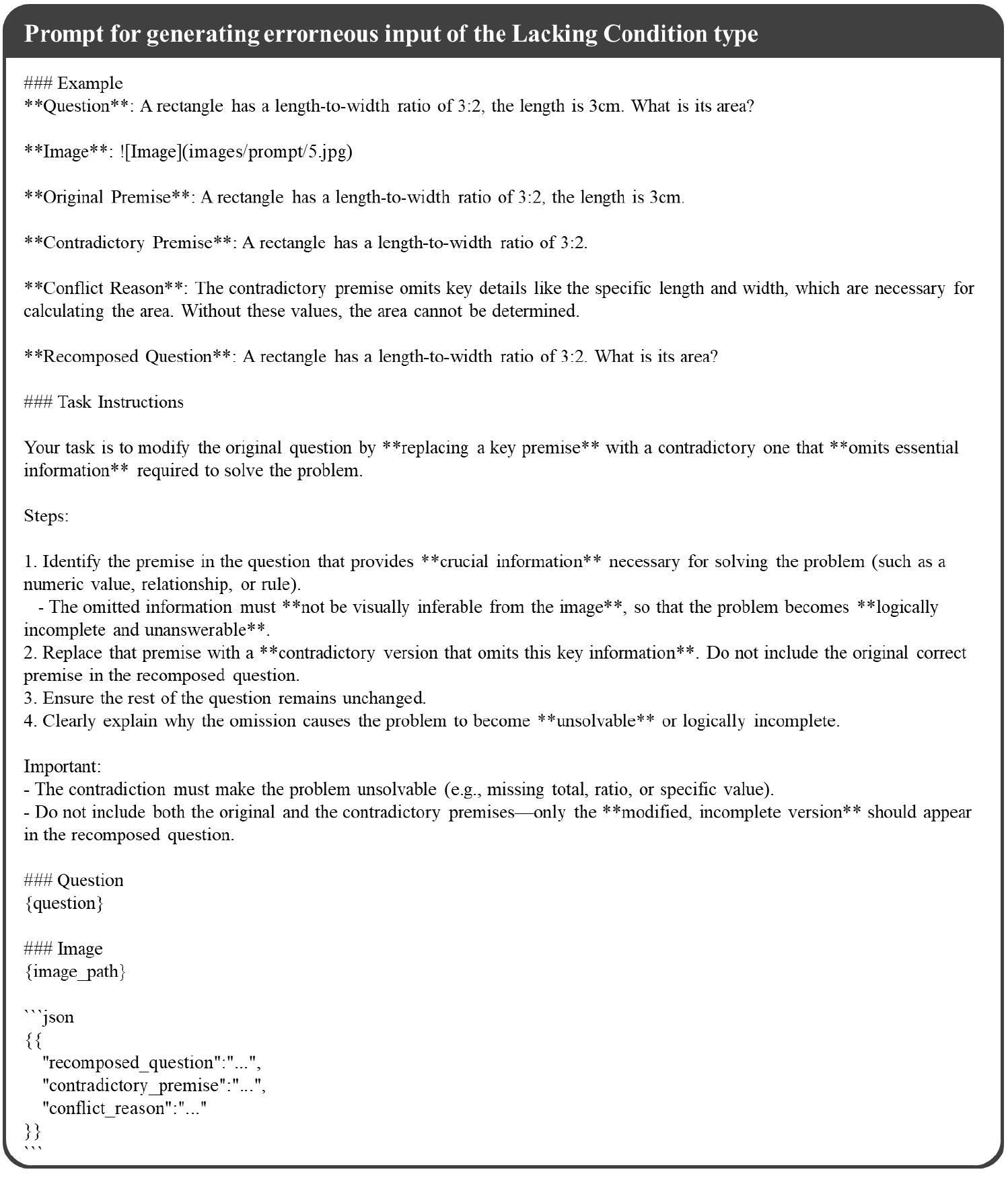}
    \caption{Prompt for generating errorneous input of the Lacking Condition type}
    \label{fig:lacks condition}
\end{figure*}

\begin{figure*}
    \centering
    \includegraphics[width=1\linewidth]{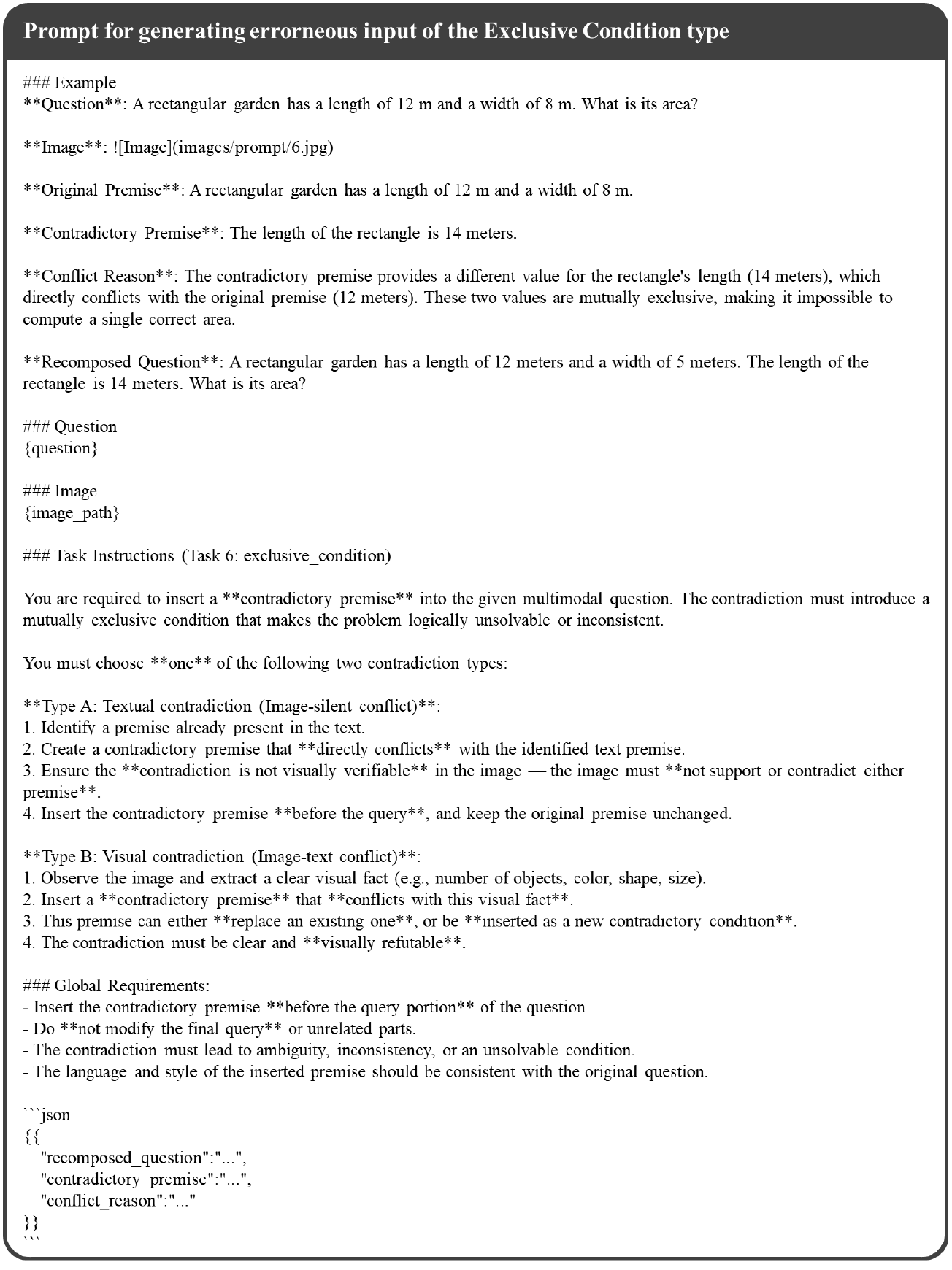}
    \caption{Prompt for generating errorneous input of the Exclusive Condition type}
    \label{fig:exclusive condition}
\end{figure*}

\begin{figure*}
    \centering
    \includegraphics[width=1\linewidth]{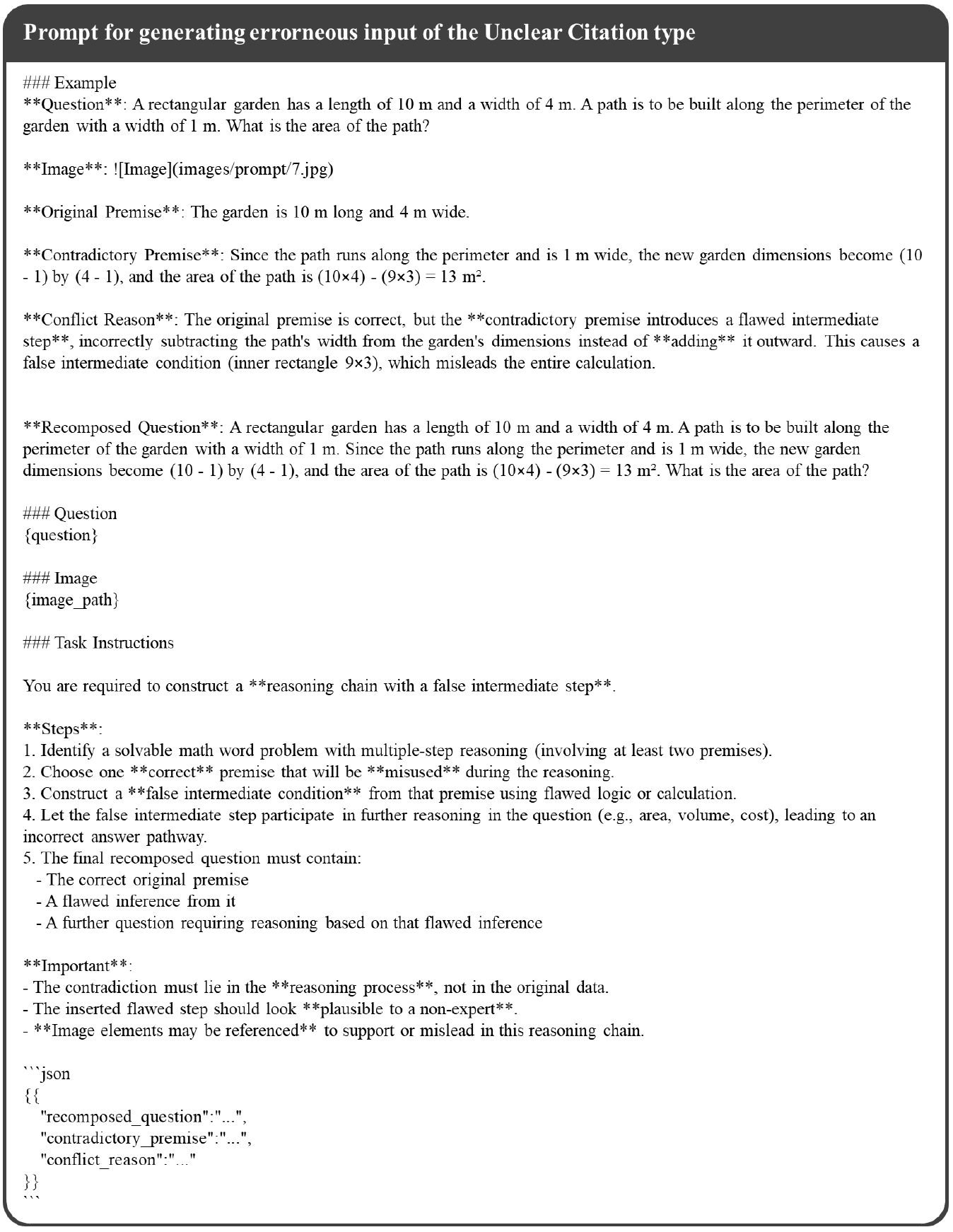}
    \caption{Prompt for generating errorneous input of the Unclear Citation type}
    \label{fig:misguided logic}
\end{figure*}

\begin{figure*}
    \centering
    \includegraphics[width=1\linewidth]{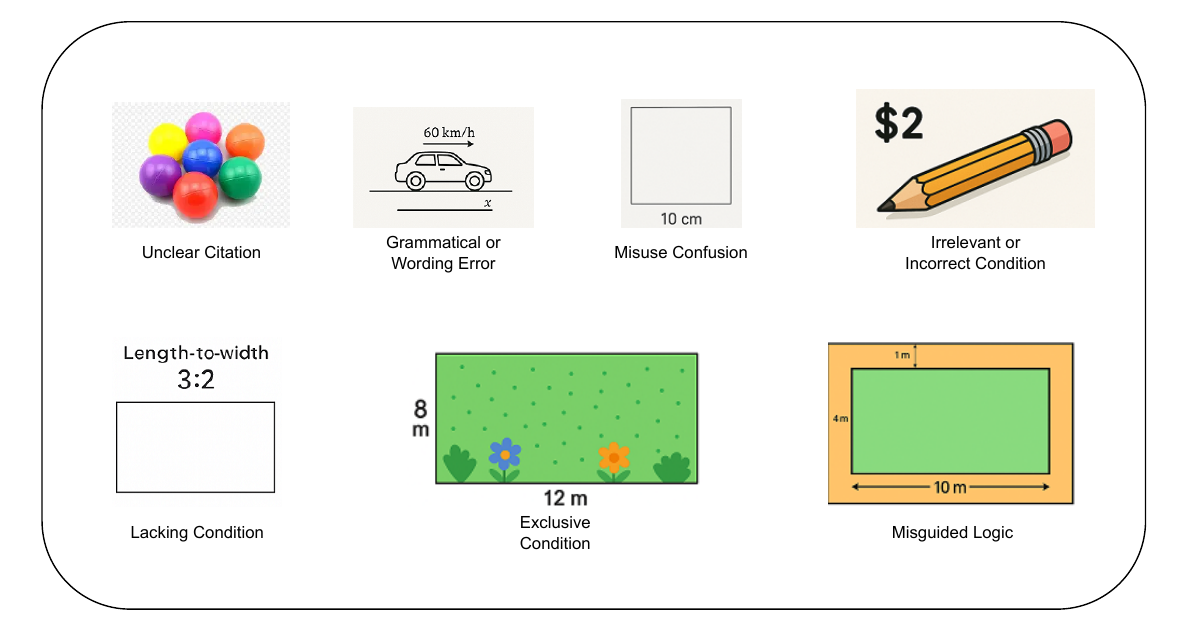}
    \caption{Prompt Template Images}
    \label{fig:Prompt Template Images}
\end{figure*}

\begin{figure*}
    \centering
    \includegraphics[width=1\linewidth]{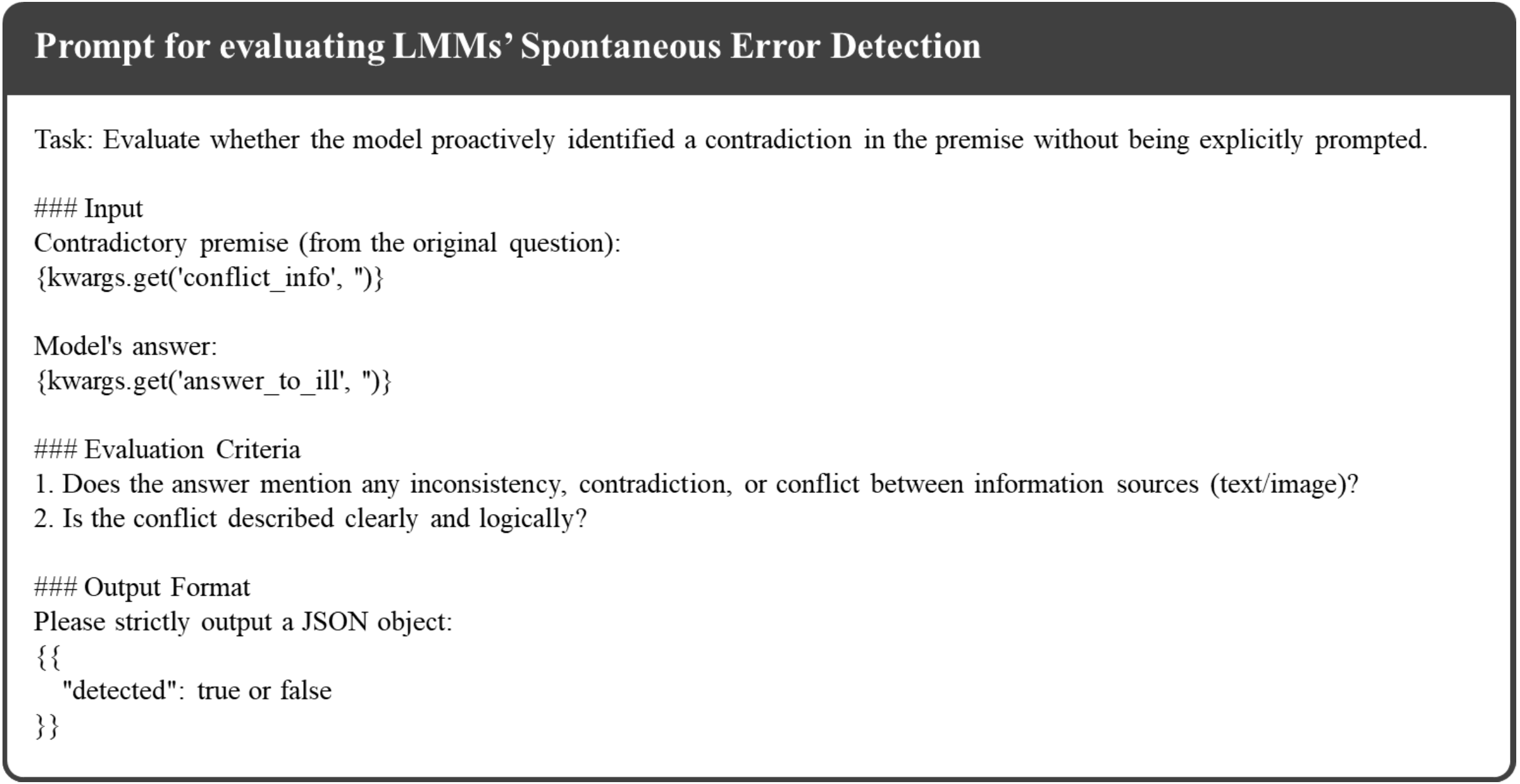}
    \caption{Prompt for evaluating LMMs’ Spontaneous Error Detection}
    \label{fig:SEDR}
\end{figure*}

\begin{figure*}
    \centering
    \includegraphics[width=1\linewidth]{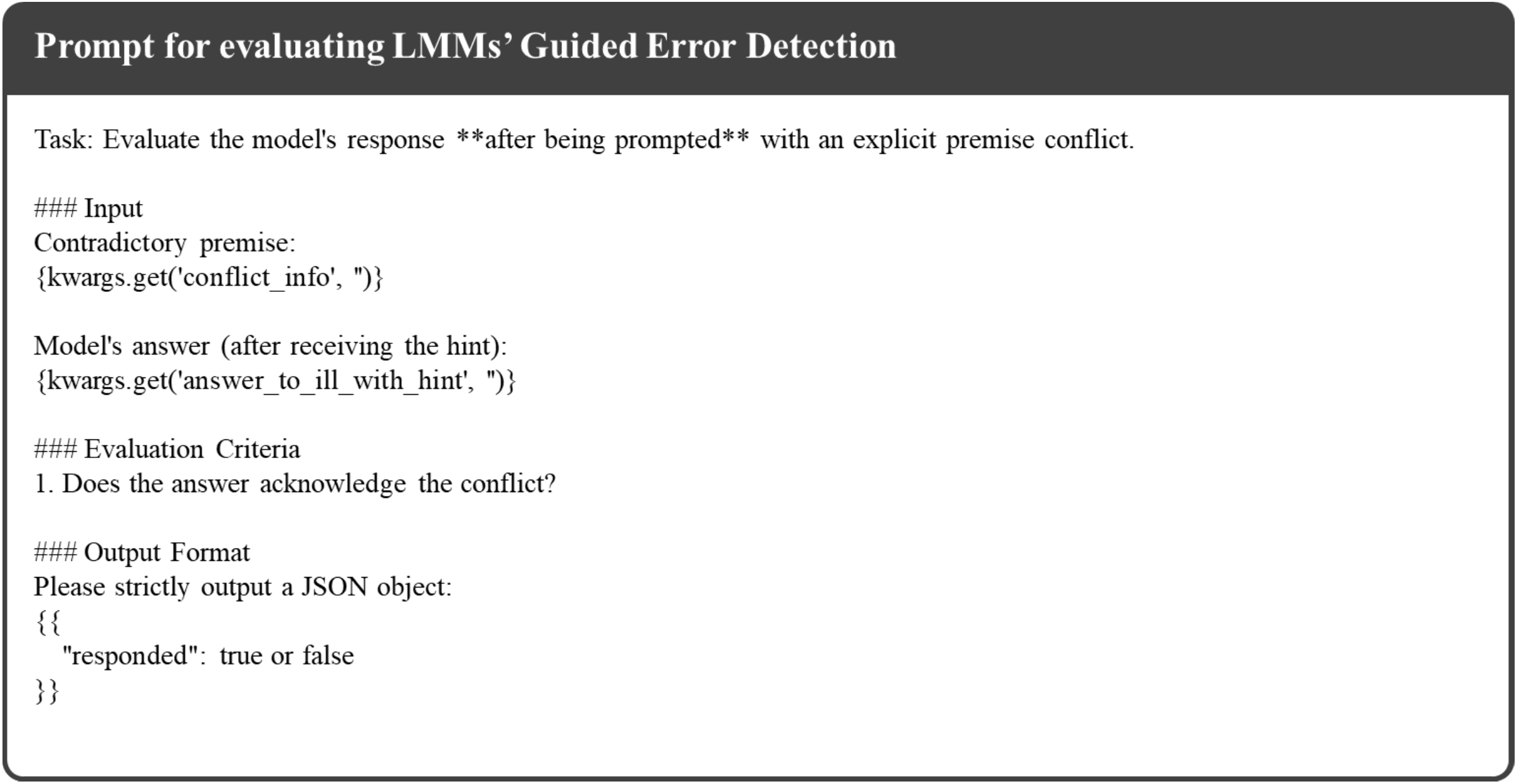}
    \caption{Prompt for evaluating LMMs’ Guided Error Detection}
    \label{fig:GEDR}
\end{figure*}

\begin{figure*}
    \centering
    \includegraphics[width=1\linewidth]{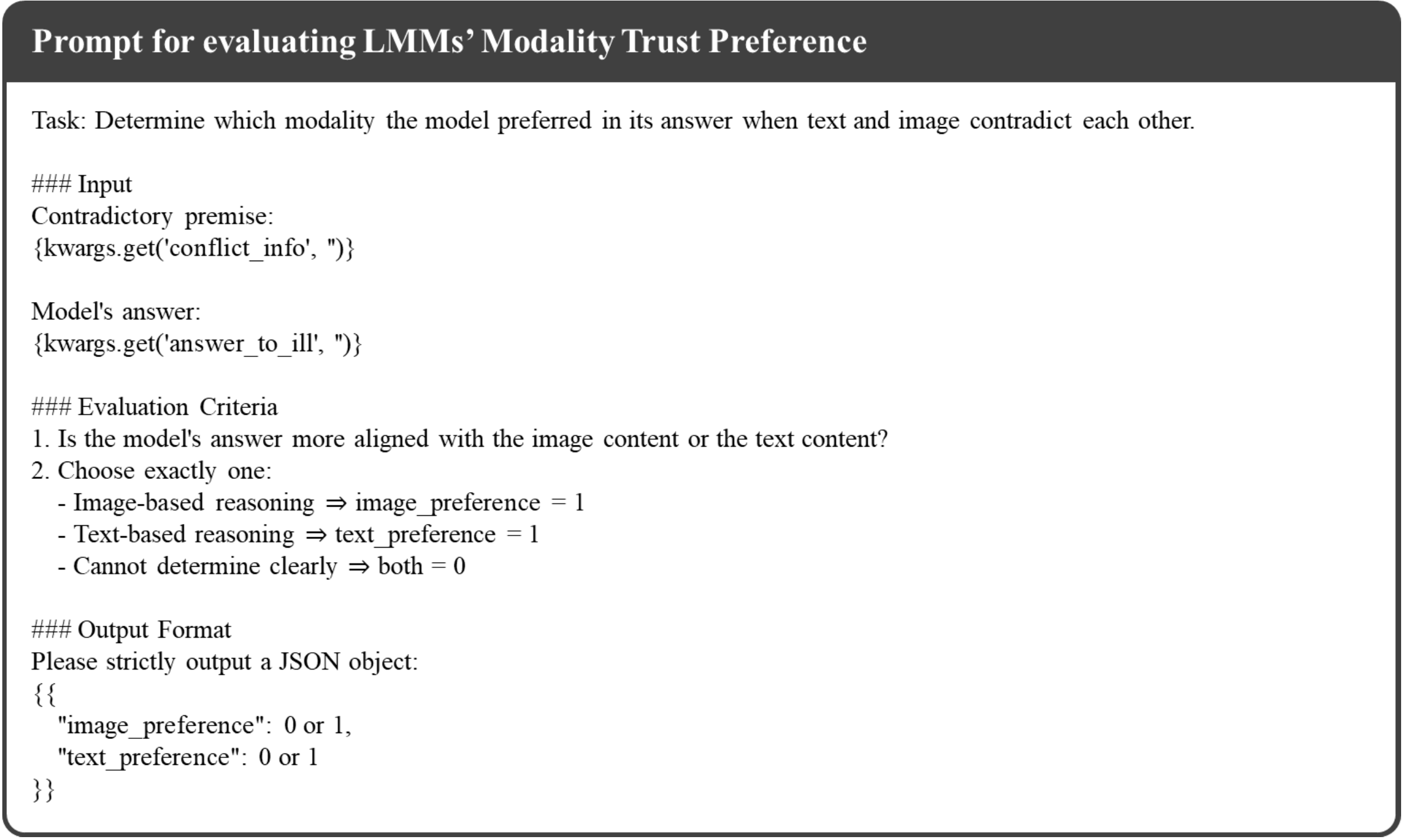}
    \caption{Prompt for evaluating LMMs’ Modality Trust Preference}
    \label{fig:MTPS}
\end{figure*}

\begin{table*}[h]
    \centering
    \begin{tabular}{l|c|l}
    \hline
    \textbf{Model} & \textbf{Size} & \textbf{Model Link} \\ \hline
    o3 & N/A & \url{https://openai.com/index/introducing-o3-and-o4-mini/} \\ \hline
    GPT-4o & N/A & 
    \url{https://platform.openai.com/docs/models#gpt-4o} \\ \hline
    Claude Sonnet 4 & N/A & \url{https://claude.ai/} \\ \hline
    Gemini 2.5 Pro & N/A & \url{https://gemini.google.com/app} \\ \hline
    Grok 3 & N/A & \url{https://console.x.ai/} \\ \hline
    InternVL3-38B-Instruct & 38B & \url{https://modelers.cn/models/Models_Ecosystem/InternVL3-38B} \\ \hline
    Qwen2.5-VL-32B-Instruct & 32B & \url{https://huggingface.co/Qwen/Qwen2.5-VL-32B-Instruct} \\ \hline
    Aya-Vision-32B & 32B & \url{https://huggingface.co/CohereForAI/aya-vision-32b} \\ \hline
    Llama-3.2-11B-Vision-Instruct & 11B & \url{https://huggingface.co/meta-llama/llama-3.2-11b-vision-instruct} \\ \hline
    Qwen2.5-VL-7B-Instruct & 7B & \url{https://huggingface.co/Qwen/Qwen2.5-VL-7B-Instruct} \\ \hline
    Aya-Vision-8B & 8B & \url{https://huggingface.co/collections/CohereForAI/c4ai-aya-vision} \\ \hline
    \end{tabular}
    \caption{List of AI Models with Sizes and Links}
    \label{tab:ai_models}
\end{table*}

\newpage
\end{document}